\documentclass[letterpaper]{article} 
\usepackage{aaai24}  
\usepackage{times}  
\usepackage{helvet}  
\usepackage{courier}  
\usepackage[hyphens]{url}  
\usepackage{graphicx} 
\usepackage{natbib}  
\usepackage{caption} 
\usepackage{algorithm}

\usepackage{newfloat}
\usepackage{listings}
\DeclareCaptionStyle{ruled}{labelfont=normalfont,labelsep=colon,strut=off} 
\floatstyle{ruled}
\newfloat{listing}{tb}{lst}{}
\floatname{listing}{Listing}
\usepackage{amsmath}
\usepackage{amssymb}

\usepackage{algpseudocode}
\usepackage{array}
\usepackage{tablefootnote}
\usepackage{dsfont}
\usepackage{booktabs}
\usepackage{cleveref}
\usepackage{mathtools}

\newcommand{\REAL} {\mathbb{R}}

\newcommand{\PHINEIGHBORHOOD} {$\phi_{0}$-neighborhood }

\usepackage[utf8]{inputenc}

\DeclareMathSizes{6.5pt}{6.5pt}{6.5pt}{6.5pt} 
\expandafter\patchcmd\csname\string\algorithmic\endcsname%
      {\labelwidth 0.5em}{\labelwidth0pt\labelsep0pt}{}{}

\title{Active Reinforcement Learning for Robust Building Control}
\author {
    Doseok Jang,
    Larry Yan,
    Lucas Spangher, 
    Costas Spanos
}
\affiliations {
    UC Berkeley\\
    djang@berkeley.edu, yanlarry@berkeley.edu, lucas\_spangher@berkeley.edu, spanos@berkeley.edu
}

\begin{document}

\maketitle

\begin{abstract}
  Reinforcement learning (RL) is a powerful tool for optimal control that has found great success in Atari games, the game of Go, robotic control, and building optimization. RL is also very brittle; agents often overfit to their training environment and fail to generalize to new settings. Unsupervised environment design (UED) has been proposed as a solution to this problem, in which the agent trains in environments that have been specially selected to help it learn.  Previous UED algorithms focus on trying to train an RL agent that generalizes across a large distribution of environments. This is not necessarily desirable when we wish to prioritize performance in one environment over others. In this work, we will be examining the setting of robust RL building control, where we wish to train an RL agent that prioritizes performing well in normal weather while still being robust to extreme weather conditions. We demonstrate a novel UED algorithm, ActivePLR, that uses uncertainty-aware neural network architectures to generate new training environments at the limit of the RL agent's ability while being able to prioritize performance in a desired base environment. We show that ActivePLR is able to outperform state-of-the-art UED algorithms in minimizing energy usage while maximizing occupant comfort in the setting of building control.
\end{abstract}

\section{Introduction}

Reinforcement learning has demonstrated remarkable success in solving sequential decision-making tasks such as the game of Go \cite{silver2017mastering}, Atari games \cite{mnih2013playing}, energy pricing \cite{jang2021offline, gunn2022adversarial}, and many others. However, RL agents often overfit to their training environment and fail to generalize to new environments\cite{zhang2018study}. This is a serious issue in tasks where we expect underlying dynamics in the environment not to stay static, e.g. when there is distribution shift between the training environment and the test environment. Here, we explore the use of RL in residential and commercial building control, where most often an agent is trained to optimize performance in normal weather conditions. When underlying weather conditions exhibit extremes or drift due to long term effects such as climate change, control often fails. We endeavor to address distribution shift by using uncertainty to select training environments such that the resulting RL agent performs well in average conditions and is also robust to uncommon but dangerous scenarios in the test environment.

\subsection{Overview of HVAC Setpoint Control}
Our focus is on robust RL for building energy consumption, which represent $73\%$ of electricity usage and $40\%$ of greenhouse gases in the US \cite{DOE}. Buildings are generating increasingly large amounts of sensory information that can be used to increase energy efficiency, such as temperature, airflow, humidity, occupancy, light, and energy usage \cite{hayat2019state}.

There are several ways buildings can be automatically controlled: heating, ventilation, and air conditioning (HVAC) units, energy storage systems, plug-in electric vehicles, photovoltaic power sources, and lights \cite{gong2022comprehensive}. We will focus on HVAC as it represents roughly a third of total building energy consumption \cite{wemhoff2010predictions}. Traditionally, HVAC setpoint control has been approached through model-predictive control (MPC, \citealp{Kou2021ModelBasedAD}) or a heuristic rule-based-controller (RBC, \citealp{Mathews2001HVACCS}).\footnote{"setpoints" are the numbers a human might input into their thermostat to tell the HVAC systems their desired temperature. "Setpoint control" is the problem of automatically determining these setpoints to optimize some objective.} MPC is generally not scalable to high dimensional input or output spaces compared to RL, and heuristics are inflexible.

\begin{figure}[t]
    \centering
    \includegraphics[width=0.4\textwidth]{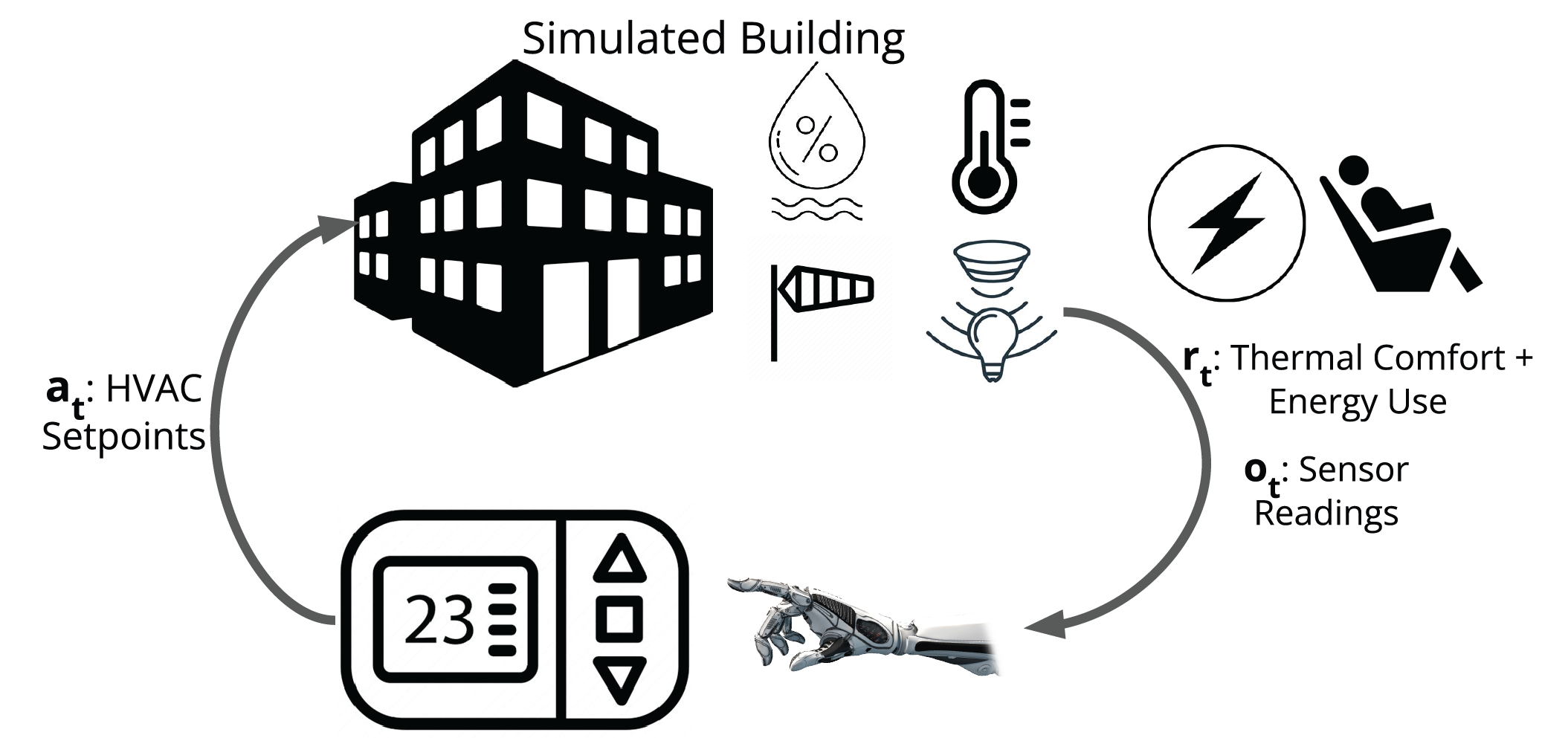}
    \caption{HVAC setpoint control in Sinergym. A simulated building sends sensor data as observations to an RL agent, which responds with HVAC setpoints as actions, and is rewarded according to energy use and thermal comfort. }
    \label{fig:sinergym}
\end{figure}

Recently, RL-based HVAC setpoint control has grown in popularity\cite{das2022machine}. \citet{rizvi2022experimental} used Q-learning to control HVAC setpoints in the presence of unseen disturbances. \citet{Kurte2020EvaluatingTA} demonstrated how RL and meta-RL could train an RL HVAC agent that quickly adapts to different buildings.  \citet{xu2020one} explored how to use transfer learning to train an RL HVAC controller that outperforms a RBC baseline across a variety of simulated buildings and climates. \Cref{fig:sinergym} illustrates the flow of information in our RL HVAC control setup.

As climate change continues, we will experience more droughts, heat waves, rising temperatures, and cold snaps \cite{masson2021climate}. These extreme weather events are likely underrepresented in the training data, but are essential to account for in reliable and safe building control. For example, consider an RL controller trained on the dry weather of the fictional country of Desertland. If the climate of Desertland changes to have increased humidity, the RL controller may react by raising temperature during a heat wave, which is energy-inefficient, unacceptable for occupant comfort, and may even threaten occupant health. 

To our knowledge, the problem of training RL HVAC controllers that are robust to changing weather is understudied, and current work focuses on detecting climate change and retraining RL controllers as their performance drops over time \cite{naug2022deep, deng2022towards}, or training RL controllers that can be transferred to buildings in other climates\cite{xu2020one, lissa2020transfer}. These works focus on RL training pipelines that are robust to long-term changes in climate, but will still underperform during short-term extreme weather events that are rare in their training distribution.


\subsection{Unsupervised Environment Design}
The process of automatically selecting areas of the state space to explore is known as active learning, or as optimal experiment design. Active learning has mostly been explored for supervised learning. For example, \citet{cohn1994improving, faria2022joint} find regions of uncertainty in the data distribution through misclassification rates and output entropy. \citet{makili2012active} uses conformal prediction to quantify the similarity of new data points to their dataset. EVOP \cite{lynch2003evop} uses the sequential simplex method in order to identify experiment configurations that can maximize information gain. \citet{bouneffouf2016exponentiated} use random exploration to identify new, promising data samples. Some of these concepts are already in use in many RL algorithms; for example, the RL algorithm we use in this paper (PPO \citealp{schulman2017proximal}), is incentivized to explore new data samples via random exploration and increasing the output action distribution entropy.

Our setting of Unsupervised Environment Design (UED, \citealp{dennis2020emergent}) is related, but different from active learning, in that we are not directly selecting training data \textit{points} to sample, but selecting \textit{parameters} of the environment that generates training data points. This is a more helpful problem setting in RL, as RL agents perform well with on-policy data that is collected as the RL agent explores the environment. UED is the problem of selecting new environment parameters that maximize the RL agent's generalization across diverse environments.\footnote{Environments that can change behavior according to configuration parameters are often referred to as Procedural Content Generation environments \citep{risi2020increasing}}\citet{parker2022evolving, dennis2020emergent} find adversarial but feasible environment configurations with high regret; \citet{jiang2021prioritized} re-samples previously seen environments based on their 1-step TD error. These algorithms focus on training an RL agent that performs well across a distribution of similar tasks. SAMPLR \cite{jiang2022grounding}, a recent method, introduces the concept of curriculum induced covariate shift (CICS) and addresses it by launching several child simulations at each step to explore other step trajectories, but this approach does not scale computationally, especially when in an area such as building control, initializing state of the art building physics simulations comprises a large proportion of the overall computation.

\subsection{Contributions}
We present a novel gradient-based algorithm for UED in building control called ActivePLR that leverages agent uncertainty. To the best of our knowledge, this is the first time neural network uncertainty has been incorporated into the problem of UED; most current works focus on some form of regret; this is also the first time environment configuration variables have been directly optimized under gradient ascent rather than through some evolutionary process \cite{parker2022evolving}, resampling procedure \cite{jiang2021prioritized}, or training a separate teacher network to select new environments \cite{dennis2020emergent}. To the best of our knowledge, we are also the first to focus on training RL HVAC agents that are robust to short-term extreme weather events rather than long-term climate change.

We demonstrate how ActivePLR trains RL HVAC controllers that are (1) more performant overall, (2) robust to extreme weather conditions, and (3) more robust to the Sim2Real transfer than the current state-of-the-art in UED.

\section{Methods}
\subsection{Reinforcement Learning (RL)}
RL is a framework for finding the optimal policy in an environment. Environments are formalized by a Markov Decision Process (MDP), which consists of a tuple $(S, A, T, R)$. State and Action spaces ($S$ and $A$) consist of tuples of some fixed length indexed per timestep $t$; $T \colon S \times A \to S$ is a transition function; and $R \colon S \times A \times S \to \mathbb{R}$ is a reward function. Agents choose actions according to a probability distribution $p_{\pi_\theta}$, determined by a policy $\pi_{\theta}$ with parameters $\theta$ to optimize the RL objective $J(\theta)$, defined $J(\theta) = \mathds{E}_{\pi} \left[ \sum_{s_t,a_t \sim p_{\pi}}[r(s_t, a_t)] \right]$. (See \citealp{sutton2018reinforcement}.)

We use the PPO \citep{schulman2017proximal} RL algorithm due to its performance and previous use in building control. Many works \cite{chen2019gnu, zhang2021grid} have used PPO for HVAC control. Although we focus on PPO in this paper, our algorithm should easily extend to any actor-critic RL algorithm.


\begin{figure}[t]
    \centering
    \includegraphics[width=0.47\textwidth]{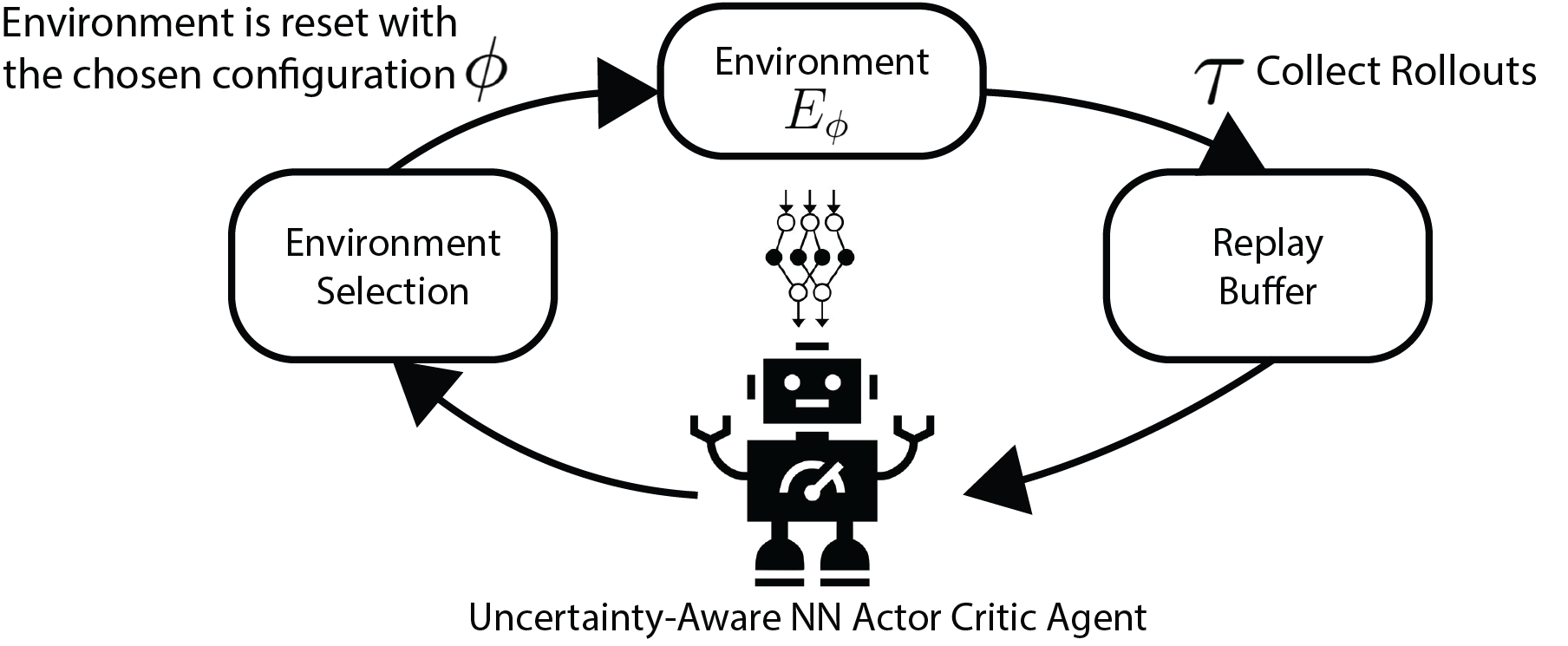}
    \caption{The flow of data during ActivePLR training.}
    \label{fig:flowchart1}
\end{figure}

\subsection{Uncertainty Estimation}
To estimate the uncertainty of our RL agent, we use Monte Carlo Dropout \cite{gal2016dropout}, in which nodes in the neural network are set to zero (``dropped out'') at random. This is used at inference time to generate multiple predictions for an individual input from different variations of the same model. The variance in these predictions is then used as a measure of the model's uncertainty. We use Monte Carlo Dropout as opposed to other methods of estimating neural network uncertainty such as bootstrapped ensembles \cite{lakshminarayanan2017simple} or Bayesian neural networks \cite{wang2020bayesian} because Monte Carlo Dropout is simpler and cheaper to train than bootstrapped ensembles and Bayesian neural networks while still providing good quantifications of uncertainty.

Formally, suppose we have a neural network $f_\theta \coloneqq \REAL^n \rightarrow \REAL^m$ mapping $n$-dimensional input vectors to $m$-dimensional output vectors. $f$ is parameterized by a list of $l$ weight matrices $\theta\coloneqq {W_i |}_{i=1}^{l}$, where $W_i \in \REAL^{N_{i-1} \times N_{i}}$ denotes the weight matrix for layer $i$ of the neural network and $N_{i}$ denotes the number of neurons in that layer. We denote the dropout operation by $d_p\colon \REAL^k \rightarrow \REAL^k$, where $d_p(\theta)$ zeros-out each column of $W_i$ in $\theta$ with probability $p$. We define the uncertainty $L$ for input $x \in \REAL^n$ as $L(x, \theta) = \operatorname{Var}(f_{d_p(\theta)}(x))$, estimated by
\begin{multline}
        L(x, \theta) = \frac{1}{C} \sum_{c=1}^{C}f_{d_p(\theta)}(x)^T f_{d_p(\theta)}(x) - \\ \mathbb{E}\left[f_{d_p(\theta)}(x)\right]^T \mathbb{E}\left[f_{d_p(\theta)}(x)\right]
\end{multline}

Essentially, we conduct $C$ independent stochastic forward passes through the model with dropout at inference time, and use the sample variance of the outputs as our uncertainty metric. To estimate the uncertainty of our RL agent, we use the uncertainty of the critic network similar to other works \cite{an2021uncertainty, wu2021uncertainty}.

\begin{figure*}[t]
    \centering
    \includegraphics[width=0.65\textwidth]{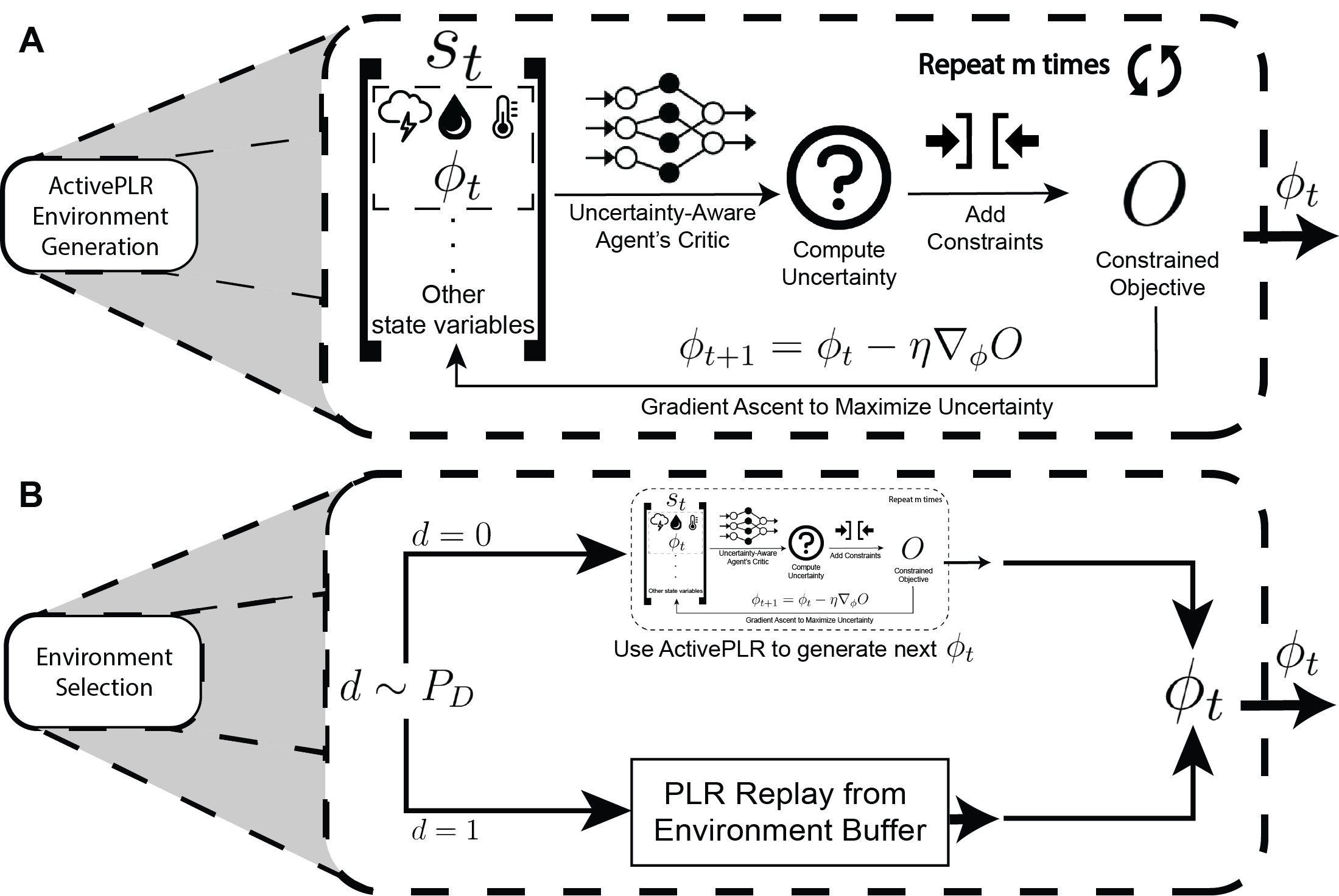}
    \caption{\textbf{A.} The ActivePLR environment generation process. \textbf{B.} The overall ActivePLR environment sampling process.}
    \label{fig:flowchart2}
\end{figure*}

\subsection{Robust Prioritized Level Replay}
\label{sec:plr}
One key assumption we make in the design of this algorithm is that, at the beginning of each training episode, we interact with an RL environment that can change its dynamics in response to some configuration parameters; for example, in this work we focus on a building simulation environment that can change its simulated weather patterns in response to weather configuration variables that we provide at the beginning of each training episode.\footnote{These environments are also known as Procedural Content Generation (PCG \cite{risi2020increasing}) environments.} Prioritized Level Replay (PLR, \citealp{jiang2021prioritized}) is a state-of-the-art framework for selectively sampling training levels in environments with procedurally generated content. Levels with higher value loss are prioritized, inducing an emergent curriculum of increasingly difficult levels.\footnote{In our case, a "level" is just an environment configuration $\phi$. We use the term "level" in describing PLR to be consistent with the original paper \cite{jiang2021prioritized}.} For each episode, PLR samples $d \sim P_D$, to decide whether to sample a new level from the training distribution $\Lambda_{\text{train}}$ or pick one from the replay buffer $\Lambda_{\text{seen}}$. \citet{jiang2021prioritized} parameterized $P_D$ as a Bernoulli distribution with probability $p = \frac{|\Lambda_{\text{seen}}|}{|\Lambda_{\text{train}}|}$. Since we consider the setting where $\phi$ is continuous, $|\Lambda_{\text{train}}|$ is infinite, so we set the denominator as a hyperparameter $N_{PLR}$, so $p = \frac{|\Lambda_{\text{seen}}|}{N_{PLR}}$.

The probability of each level in the replay buffer being sampled is determined by the value loss, and how stale that estimate of the value loss is. If we do not sample a level from the replay buffer, we sample a new one from the training distribution and add it to the buffer. In this paper, vanilla PLR samples new levels to add to the buffer uniformly at random from the set of all possible training environments. Recently, \citet{jiang2021replay} proposed Robust PLR, in which only training on the PLR-selected levels, and stopping gradient updates from the randomly selected levels, generally performs better; we will refer to this variant as RPLR.


\subsection{ActivePLR}
\label{sec:activeplr}
 We present ActivePLR: a novel addition to PLR that samples new levels to add to the replay buffer through an uncertainty-based optimization procedure instead of at random.\footnote{We also tried to use RPLR with our uncertainty-based approach, but we found it performs worse than ActivePLR} In order to generate new environments at the frontier of the agent's uncertainty, we will backpropagate gradients from the uncertainty back to the state variable and do gradient ascent.\footnote{We make the assumption that at least some of the environment configuration variables are continuous} This will allow us to find the state at which the agent is most uncertain, and generate a new environment that allows the agent to interact with the world at that state. Formally, assume we have a environment $E$ with some parameters $\phi \in \REAL^k$ that are part of the agent's initial state space $S$. That is, the state $s_0 \in \REAL^n$ can be divided into $\phi$ and $\overline{s_0}$, where $\overline{s_0} \in \REAL^{n-k}$ is defined such that $s_0 = [\phi, \overline{s_0}]$ (and $[ ]$ is the concatenation operator). We can define an objective function that tries to maximize the uncertainty of the RL agent:
\begin{equation}
    O(\phi_i, \overline{s_0}, \theta) = L([\phi_i, \overline{s_0}], \theta)
\end{equation} where $L$ is our uncertainty estimate. We can then update $\phi$:
\begin{equation}
    \phi_{i+1} = \phi_i + \eta \nabla_{\phi_i} O(\phi_i, \overline{s_0}, \theta)
\end{equation}

We use this optimization procedure to identify novel training environments to add to PLR's environment replay buffer. We call this procedure ActivePLR, as it is an active learning method that seeks to identify what data would be most useful for the RL agent.\footnote{Pseudocode for ActivePLR can be found in \Cref{alg:activeplr}, and an illustration can be found in \Cref{fig:flowchart2}.} We can also use this optimization procedure to generate all the training environments instead of using a replay buffer.\footnote{This is equivalent to ActivePLR with $P_D$ assigning $100\%$ probability to $d=0$} We denote this case as ActiveRL.

Trying to identify parameters $\phi$ by maximizing uncertainty can lead to unrealistic parameters that are outside of the test distribution and not useful for learning. Thus, we integrate both hard constraints and soft constraints on $\phi$ generation in ActivePLR. The hard constraints are useful when trying to train an RL agent that generalizes over a certain region of the $\phi$ space, and the soft constraints are useful when trying to train an RL agent that emphasizes performance near a particular $\phi_0$, which is still robust to different values of $\phi$. We will refer to the latter setting as the \PHINEIGHBORHOOD setting for brevity. 

The hard constraints constrain the search space within some lower and upper limits specified by the user for $\phi$ using the extragradient \cite{korpelevich1976extragradient} method. Suppose we have a lower bound constraint $\phi > b$ for some $b \in \REAL^k$ and an upper bound constraint $\phi < a$ for some $a \in \REAL^k$. Then we can use Lagrangian optimization methods from the Cooper library \cite{gallegoPosada2022cooper} to search for a $\phi$ with high uncertainty within the bounds of $a$ and $b$, helping to avoid unrealistic values of $\phi$.\footnote{A more detailed treatment of the constrained optimization problem can be found in \Cref{sec:hard_constraints}.} Throughout this paper, we will use the implementation of ExtragradientAdam from \citet{gallegoPosada2022cooper}, which adjusts the learning rate $\eta$ for each parameter according to Adam \cite{kingma2014adam}. 

In the \PHINEIGHBORHOOD setting, the hard constraints are not enough -- there is no guarantee that states near $\phi_0$ will be sampled. Thus we introduce a soft constraint to $O$ to minimize the Euclidean distance from $\phi$ to $\phi_0$.
\begin{equation}
    \label{eq:soft}
    O(\phi_i, s_0, \theta) = L([\phi, \overline{s_0}], \theta) - \gamma ||(\phi-\phi_0)||_2
\end{equation} where $\gamma$ emphasizes the soft constraint.
\begin{algorithm}[h]

\caption{ActivePLR}
\label{alg:activeplr}
\begin{algorithmic}
\Procedure{ActivePLR}{$\theta, s_0, N, T, \eta, \gamma, a, b, \rho, N_PLR$}\\
\Comment{$\theta\colon$policy parameters}
\Comment{$s_0\colon$initial state to seed environment generation}
\Comment{$T\colon$\# of iterations to run PPO}
\Comment{$N\colon$\# of iterations to optimize $\phi$}
\Comment{$\eta\colon$Learning rate for optimizing $\phi$}
\Comment{$\gamma\colon$Weight on soft constraint}
\Comment{$a\colon$$\phi$ lower bounds}
\Comment{$b\colon$$\phi$ upper bounds}
\Comment{$c\colon$Global episode counter}
\Comment{$\Lambda_{seen}\colon$Visited levels}
\Comment{$S\colon$Global level scores}
\Comment{$C\colon$Global level timestamps (when they were last sampled)}
\Comment{$\rho\colon$PLR staleness weighting}
\Comment{$N_{PLR}\colon$hyperparameter of $P_D$}

\State $\phi_0 \leftarrow$ ExtractPhi($s_0$)
\State $c \leftarrow c+1$
\For{t=0 to T}
    \State Sample replay decision $d \sim P_D(N_{PLR})$
    \If{$d == 1$}
        \State ScoreProb $\leftarrow P_S(\phi | \Lambda_{seen}, S)$
        \State StaleProb $\leftarrow P_C(\phi | \Lambda_{seen}, C, c)$
        \State Sample $\phi \sim (1-\rho) \cdot \text{ScoreProb} + \rho \cdot \text{StaleProb}$
    \Else
        \For{i=0 to N}
            \State $\phi \leftarrow$ ExtractPhi($s_0$)
            \State dist $\leftarrow ||\phi - \phi_0||_2$
            \State $O \leftarrow \text{UncertaintyEstimate}(f_{\theta}, s_0) - \gamma \cdot \text{dist} $
            \State $\phi \leftarrow \text{ExtragradientUpdate}(\phi, O, a, b)$
            \State $s_0$ = Concatenate([$\phi$, $\overline{s_0}$])
        \EndFor
    \EndIf
    \State Define new index $i \leftarrow |S| + 1$
    \State Add $\phi_i \leftarrow \phi$ to $\Lambda_{seen}$
    \State Add initial value $S_i=0$ to $S$ and $C_i=0$ to $C$
    \State $\tau \leftarrow$ CollectTrajectories($E_{\phi}$)
    \State Update score $S_i \leftarrow \text{PPOValueLoss}(\tau, \theta)$ 
    \State Update timestamp $C_i \leftarrow c$
    \State $\theta \leftarrow$ PPOUpdate($\tau, \theta$)
\EndFor
\State Return $\theta$
\EndProcedure
\end{algorithmic}
\end{algorithm}

\section{Experiments}

\subsection{Environment}
We use a modified version of the Sinergym \cite{2021sinergym} OpenAI Gym \cite{brockman2016openai} environment to simulate buildings in different weather conditions. The environment uses the EnergyPlus \cite{crawley2001energyplus} simulation engine to model the dynamics of building systems. We use the "5Zone" building provided with Sinergym: a $463.6 m^2$ single-story building with a DX cooling coil and gas heating coils that is divided into 5 zones (1 indoor and 4 outdoor). Actions in Sinergym are continuous, two-dimensional vectors, where the agent can control the heating and cooling setpoints for the HVAC systems. We use a reward function that rewards high occupant comfort and low energy use. To quantify occupant comfort, we use the Fanger Percentage of People Dissatisfied (PPD, \citealp{fanger1967calculation}). Energy use is the total HVAC electricity demand rate in Watts (W). The reward can be formulated as:
\begin{multline}
    R_t = - \rho * \lambda_E * P_t - \\ (1 - \rho) * \lambda_P * PPD_t * \mathds{1}_{(occupancy_t > 0)} * \mathds{1}_{PPD_t > 20}
\end{multline} where $\rho$ controls how much to weight comfort against energy use, $P_t$ is the electricity demand rate, $\lambda_E$ and $\lambda_P$ are scaling factors to account for varying units, $\mathds{1}_{(occupancy_t > 0)}$ ensures there is no penalty for uncomfortable conditions when there are no occupants, and $\mathds{1}_{PPD_t > 20}$ ensures the agent is not penalized if the PPD is below the ASHRAE guidelines' comfort threshold of $20\%$ \cite{ansi2017standard}. We use $\lambda_E=0.0001$, $\lambda_P=0.1$, $\rho=0.5$. The state is a continuous, 20 dimensional vector that includes 5 outdoor weather variables: outdoor air temperature, outdoor relative humidity, wind speed, wind direction, solar irradiance, and 15 other variables: indoor air temperature, indoor relative humidity, clothing value, thermal comfort, current HVAC setpoints, total HVAC electricity demand rate, occupancy count, and date (year, month, day, hour). 

To simulate outdoor weather, Sinergym takes as input a file with hourly measurements of each outdoor weather variable. Originally, Sinergym added noise to outdoor temperature through an Ornstein-Uhlenbeck (OU, \citealp{doob1942brownian}) process, to help prevent overfitting the agent to a static weather pattern. We modified Sinergym so it could add this noise to the other outdoor weather variables as well. An OU process has three parameters: $\sigma$, $\mu$, and $\tau$. $\sigma$ controls the variance of the added noise, $\mu$ is the average value of the noise, and $\tau$ determines how quickly the noise reverts to the mean.

We can obtain reasonable values for $\sigma$ and $\tau$ for each weather variable from the original input weather file, so we have 5 remaining parameters to customize Sinergym: the $\mu$ offset parameters for each weather variable.\footnote{See \Cref{sec:ou_details} for details} Thus the $\phi$ for Sinergym that we vary to attempt to train a robust RL agent, is the 5 dimensional vector $<\mu_1, \mu_2, \mu_3, \mu_4, \mu_5>$. These essentially change the average outdoor temperature, relative humidity, wind speed, wind direction, and solar irradiance over the course of the simulation. Varying the environment configuration $\phi \in \REAL^5$ enables us to collect training data from diverse outdoor weather conditions.

All experiments used 24 Intel Xeon E5-2670 CPUs.\footnote{Code is available at https://github.com/Demosthen/ActiveRL}

\subsection{Baseline Algorithms for HVAC Control}
Our basic RL baseline is a PPO agent composed of a neural network with two hidden layers, each with 256 neurons, using dropout and ReLU activations, that is trained on $\phi_0$. 

The most common method of training agents that generalize across diverse environments is domain randomization (DR), where $\phi$ is selected uniformly at random. This method is often used so agents can transfer from simulation to the real world \cite{chen2021understanding, vuong2019pick, tobin2017domain}. In the buildings domain, \citet{jang2021offline} used DR to train an RL energy pricing agent to be robust to the Sim2Real transfer. In our setting: we have some lower bounds $a \in \REAL^5$ and upper bounds $b \in \REAL^5$ for each variable, described in \Cref{tab:bounds} in the appendix\footnote{The appendix can be found}. We sample $\phi \sim U(a, b)$, where $U$ is the uniform distribution. 

\citet{jiang2022grounding} propose Sampled Matched PLR (SAMPLR) to generalize across diverse environments while combating the problem of curriculum-induced covariate shift (CICS), in which the distribution of training environments ($\phi \sim P$) generated through UED may become too different from the distribution of test environments ($\phi_{test} \sim \Bar{P}$). Unfortunately we were not able to use SAMPLR as a baseline: SAMPLR resets the simulator at every timestep to collect fictitious trajectories; thus,  SAMPLR is infeasible when a reset is expensive compared to a timestep, which is true for Sinergym and many other simulators.\footnote{Sinergym takes $\sim 3$ seconds to reset to a new weather pattern. Each of our experiments involved training for 3M timesteps, where episodes were reset every 8760 timesteps. Implementing SAMPLR would have increased the time spent resetting Sinergym by 8760x, for a total of 9M seconds, or 104 days.}


We use a rule-based controller (RBC) based on Sinergym's RBC, and a random controller as baselines. The random controller outputs a random cooling setpoint $a[0] \sim Uniform(22.5, 30.0)$ and a heating setpoint $a[1] \sim Uniform(15, 22.5)$.\footnote{The specific values are taken from Sinergym.} Sinergym's RBC sets the desired temperature range ($26$-$29\text{\textdegree} C$) higher in the summer and lower in the winter ($20$-$23.5\text{\textdegree}C$) to minimize energy consumption. We added a rule that if no occupants are in the building, the RBC sets setpoints with a wide enough range that the HVAC system is turned off. We added this new occupancy-based rule for the sake of a fair comparison because we included occupancy information in the reward.

Our last RL baseline, RPLR, is described in \Cref{sec:plr}.

Hyperparameters are in \Cref{tab:hyperparameters} in the appendix.

\subsection{Evaluating ActivePLR's Robustness to Extreme Weather Events and the Sim2Real Jump}
\label{sec:extreme_setup}
In order to evaluate how robust agents trained by each algorithm are to extreme weather events, we evaluate each agent in a suite of 5 different extreme weather environments parameterized by 5 different $\phi$ weather configurations, as well as $\phi_0$ for a total of 6 environments. The agent is trained on automatically generated environments according to each UED algorithm. It is then evaluated in the following 6 environments: $\phi_0$ simulates realistic weather based on recordings from Arizona, USA, $\phi_1$ simulates an extremely hot and dry drought, $\phi_2$ simulates a wet and windy storm, $\phi_3$ simulates a humid heatwave, $\phi_4$ simulates a cold snap, and $\phi_5$ simulates erratic weather. \textbf{Our hypothesis is that using uncertainty to identify new environments to collect data from will allow us to train RL agents that are more robust to extreme weather conditions.}

\label{sec:sim2real_setup}
In order to test whether or not the RL policies trained in simulation can be extended to the real world, we evaluated each RL algorithm in each of the 6 environments by running the EnergyPlus simulator at a higher fidelity than the agents were trained on, thus simulating the "Sim2Real" jump with a more realistic simulator. The agents were trained on a simulator operating at a granularity of $\Delta t = 1 \text{hour}$ per timestep, and we evaluate on a granularity of $\Delta t = 0.25 \text{hours}$ per timestep. During evaluation, each of the RL agents' actions are simply repeated four times so that it still takes an action every hour. \textbf{Our hypothesis is that by increasing the state space supported by the training distribution, ActivePLR will help the agent be robust to compounding errors caused by the Sim2Real jump.}

\begin{figure*}[ht!]
    \centering
    \includegraphics[width=0.7\textwidth]{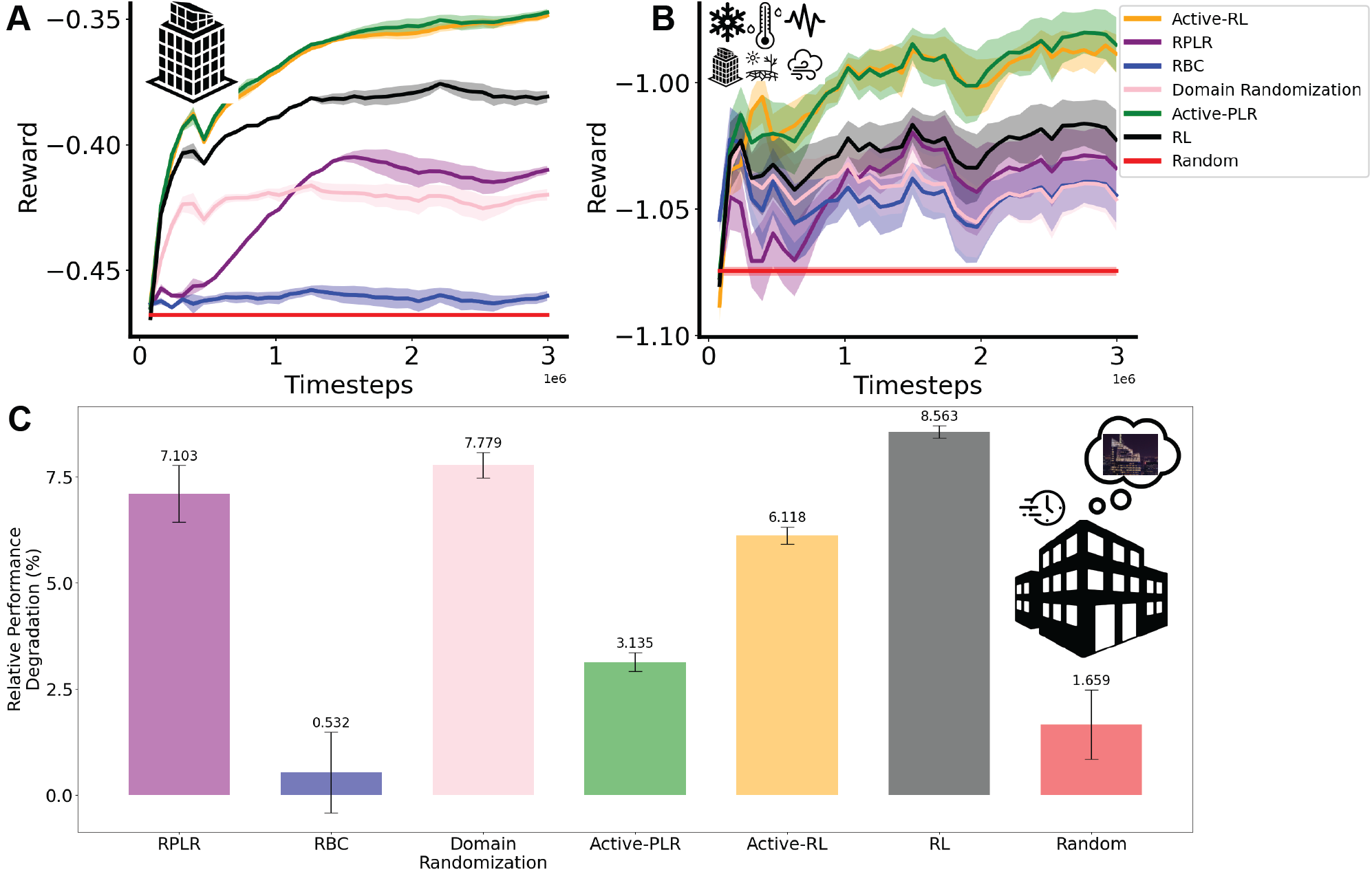}
    \caption{Performance of each algorithm on training an RL HVAC agent in Sinergym, tested on various weather patterns. ActiveRL and ActivePLR outperform all baselines. We report the standard error of the mean over 5 trials for each result. Each ActiveRL and ActivePLR trial took 12 hours to train and evaluate. \textbf{A.} Average reward of each algorithm on the base environment $\phi_0$ throughout training.  \textbf{B.} Average reward achieved by each algorithm averaged over all 6 environments throughout training. \textbf{C.} The average drop in reward over all environments when evaluating in the higher fidelity simulation compared to evaluation in the lower fidelity simulation. Lower is better here. Note that none of the algorithms were trained on the specific extreme weather environments. Evaluating ActiveRL and ActivePLR on the higher fidelity simulation took about 1 hour per trial.}
    \label{fig:results}
\end{figure*}

\section{Results and Discussion}
\subsection{ActivePLR is Robust to Extreme Weather Events}
\label{sec:extreme_results}
In order to evaluate how robust agents trained by each algorithm are to extreme weather, we evaluate the agent in a suite of 6 different environments, parameterized by 5 different $\phi$ extreme weather configurations and $\phi_0$. \Cref{fig:results} shows the overall performance of each algorithm. See \Cref{fig:results_big} in the appendix for performance in specific environments.

Surprisingly, DR and RPLR did not have significant improvements over the vanilla RL algorithm. By the end of training, DR and RPLR achieved $9\%$ higher reward than the RBC on the base environment $\phi_0$ after 3M timesteps of training. However, the vanilla RL policy had a $8\%$ improvement over the DR and RPLR policies with $\phi_0$. Over the 5 extreme weather environments, DR did about as well as the RBC. The unexpected lack of performance gain may mean the environments generated with DR were too unrealistic to learn how to perform in extreme weather conditions. 

Over the extreme weather conditions, RPLR beat DR and the RBC, but performed worse than the vanilla RL policy. This indicates that as RPLR uses DR to sample new environments, it may still suffer from generating unrealistic environments. However, its weighted environment resampling procedure helps it generalize better than naive DR, even though it is resampling from unrealistic environments. We found that training the HVAC controller with ActivePLR resulted in agents that performed better in both the extreme environments and the base environment. 

Generally, ActiveRL and ActivePLR performed similarly. In three out of the five extreme environments: (1) the hot drought, (2) the cold and windy, and (3) the cold snap environment, ActivePLR performed significantly better than all other baselines and performed competitively in the remaining two environments. In the base environment, ActivePLR provides a $9\%$ improvement over vanilla RL, and a $24\%$ improvement over RBC. Over all 6 environments, it provides a $3\%$ improvement over vanilla RL. We also found that over the 6 environments, ActivePLR and ActiveRL provided a $15\%$ relative decrease in days with ASHRAE thermal comfort violations compared to vanilla RL (which was the best baseline in terms of thermal comfort), resulting in significantly more comfortable occupants even during extreme weather conditions.\footnote{ASHRAE defines uncomfortable thermal conditions as at least $20\%$ of occupants are predicted to be uncomfortable (PPD $>20\%$)} The fact that ActivePLR significantly outperforms all baselines indicates there is considerable value in seeking out realistic new training environments that maximize an agent's uncertainty rather than choosing environments at random or merely replaying old ones.

\subsection{ActivePLR Generalizes from Simulation}
\label{sec:sim2real}
One flaw in this work and UED algorithms in general is that a simulator is required to train the model in different environments. Thus, it is important to ask whether or not the RL policies trained in simulation can be extended to the real world. In order to approximate the Sim2Real gap, we conducted evaluated each RL algorithm in each of the 6 environments by running the EnergyPlus simulator at a higher fidelity than the agents were trained on, so that these test environments (1) had slightly different dynamics from the original training simulation, which should give rise to similar distribution shift issues as the Sim2Real gap, and (2) had dynamics that were as close to those of the real world as possible because the test simulation was run with higher fidelity and should therefore be more accurate to real dynamics than the training simulation. An illustration of the performances of each algorithm on each of the 6 handcrafted environments from \Cref{sec:extreme_results} are shown in \Cref{fig:results}. 

When the agents are transferred from the simulation to our surrogate for the real world, we see there is a significant performance drop across all data-driven algorithms. Vanilla RL achieves a reward that is $8.5\%$ lower on average across the 6 handcrafted environments when evaluated on the higher fidelity simulation. DR and RPLR have smaller relative drops of about $7 \%$. However, since the $7 \%$ is relative to the performance of DR and RPLR in the original low fidelity simulation, vanilla RL still performs better in terms of absolute reward over the six environments. Random and RBC have very small or no relative performance degradation, which is to be expected because they are not data-driven models. They still perform the worst in terms of absolute reward. ActiveRL and ActivePLR, however, achieve both smaller relative drops in performance and higher absolute reward across all the different extreme weather scenarios. ActiveRL has only a $6.1\%$ relative drop in reward while ActivePLR has only a $3.1\%$ relative drop. These are promising results that indicate that these algorithms would still perform well if deployed in the real world after being trained in simulation. Furthermore, these algorithms result in agents that are more robust to the Sim2Real transfer than other methods. 

We found that ActivePLR trains agents that are more robust to the Sim2Real transfer than ActiveRL, which is surprising since ActivePLR was not significantly different from ActiveRL in the extreme weather experiment. There might be some attractive local optimum in the HVAC control task in the low fidelity simulation that both ActivePLR and ActiveRL fall into that is not present in the high fidelity simulation, resulting insimilar performance in the experiments from \Cref{sec:extreme_results} but better Sim2Real transfer. In addition, the recorded value loss that RPLR and ActivePLR use is likely a less noisy signal of environment curriculum value than the uncertainty over the value estimate that ActiveRL uses. The value loss is obtained by actually collecting data while the value uncertainty is estimated using only the model weights through Monte Carlo Dropout.

\section{Conclusion and Limitations}
We explored the utility of a novel uncertainty-driven, gradient based algorithm called ActivePLR for unsupervised environment design in the context of training RL building control agents that are robust to climate change. We found that incorporating uncertainty into UED through ActivePLR led to HVAC controllers that better optimized thermal comfort and energy usage, even in extreme weather scenarios that were never in the training distribution. Our experiments showed that other UED algorithms perform poorly when generating new environment configurations for weather patterns because they may output unrealistic weather patterns that do not help the RL agent perform well in more realistic weather scenarios. Furthermore, we showed that ActivePLR and its variant ActiveRL would have a much smaller degradation in performance when transferring from the simulated domain to the real world compared to other techniques, making them a practical option for training robust RL HVAC agents that are ready for real deployment.

This work has two primary limitations. The first is that we rely on simulations; this is a flaw that is shared by most work that focuses on UED as well as many works in building control, as access to real buildings is difficult to obtain. The second is that our method requires continuous environment configuration variables to conduct gradient ascent. Future work could explore how this could be mitigated by applying dequantization techniques \cite{das2022improved} that transform categorical variables into continuous variables.


\section{Acknowledgements}
This research is funded by the Republic of Singapore's National Research Foundation through a grant to the Berkeley Education Alliance for Research in Singapore (BEARS) for the Singapore‐Berkeley Building Efficiency and Sustainability in the Tropics (SinBerBEST) Program. BEARS has been established by the University of California, Berkeley as a center for intellectual excellence in research and education in Singapore.

\bibliography{aaai24}
\appendix
\section{ActivePLR Hard Constraint Details}
\label{sec:hard_constraints}
The hard constraints constrain the search space within some lower and upper limits specified by the user for $\phi$ using the extragradient \cite{korpelevich1976extragradient} method. Suppose we have a lower bound constraint $\phi > b$ for some $b \in \REAL^k$ and an upper bound constraint $\phi < a$ for some $a \in \REAL^k$. Then we can express the Lagrangian as 
\begin{equation}
    \begin{split}
    \mathcal{L}(\phi, s_0, \theta, \lambda, \mu) = O(\phi, s_0, \theta) + \\    \sum_{i} \lambda_i (b_i - \phi_i) + \sum_{i} \mu_i (\phi_i - a_i)
    \end{split}
\end{equation}

Now we can express the extragradient update. First, define the joint variable $\omega = (\phi, \lambda, \mu)$. We omit $s_0$ and $\theta$ from $\omega$ and the parameters to $\mathcal{L}$ as they will be kept constant throughout the optimization process. Then, the extragradient optimization process can be described as:
\begin{equation}
    F(\omega) = [\nabla_{\phi} \mathcal{L}(\omega), -\nabla_{\lambda} \mathcal{L}(\omega) - \nabla_{\mu} \mathcal{L}(\omega)]^T
\end{equation}
\begin{equation}
    \omega_{t+1/2} = P_{\Omega}[\omega_t - \eta F(\omega_t)]
\end{equation}
\begin{equation}
    \omega_{t+1} = P_{\Omega}[\omega_t - \eta F(\omega_{t+1/2})]
\end{equation} where $P_\Omega[\cdot]$ is the projection onto the constraint set.

\begin{algorithm}[h]
\caption{ActiveRL}
\label{alg:activerl}
\begin{algorithmic} 
\Procedure{ActiveRL}{$\theta, s_0, N, T, \eta, \gamma, a, b, p$}\\
\Comment{$\theta\colon$policy parameters}
\Comment{$s_0\colon$initial state to seed environment generation}
\Comment{$T\colon$number of iterations to run PPO}
\Comment{$N\colon$number of iterations to optimize $\phi$}
\Comment{$\eta\colon$Learning rate for optimizing $\phi$}
\Comment{$\gamma\colon$Weight on soft constraint}
\Comment{$a\colon$$\phi$ lower bounds}
\Comment{$b\colon$$\phi$ upper bounds}

\State $\phi_0 \leftarrow$ ExtractPhi($s_0$)
\For{t=0 to T}
    \For{i=0 to N}
        \State $\phi \leftarrow$ ExtractPhi($s_0$)
        \State $O \leftarrow \text{UncertaintyEstimate}(f_{\theta}, s_0) - \gamma ||\phi - \phi_0||^2$
        \State $\phi \leftarrow \text{ExtragradientUpdate}(\phi, O)$
        \State $s_0$ = Concatenate([$\phi$, $\overline{s_0}$])
    \EndFor
    \State $\tau \leftarrow$ PPOCollectTrajectories($E_{\phi}$)
    \State $\theta \leftarrow$ PPOUpdate($\tau, \theta$)
\EndFor
\State Return $\theta$
\EndProcedure
\end{algorithmic}
\end{algorithm}

\begin{figure*}
    \centering
    \includegraphics[width=0.8\textwidth]{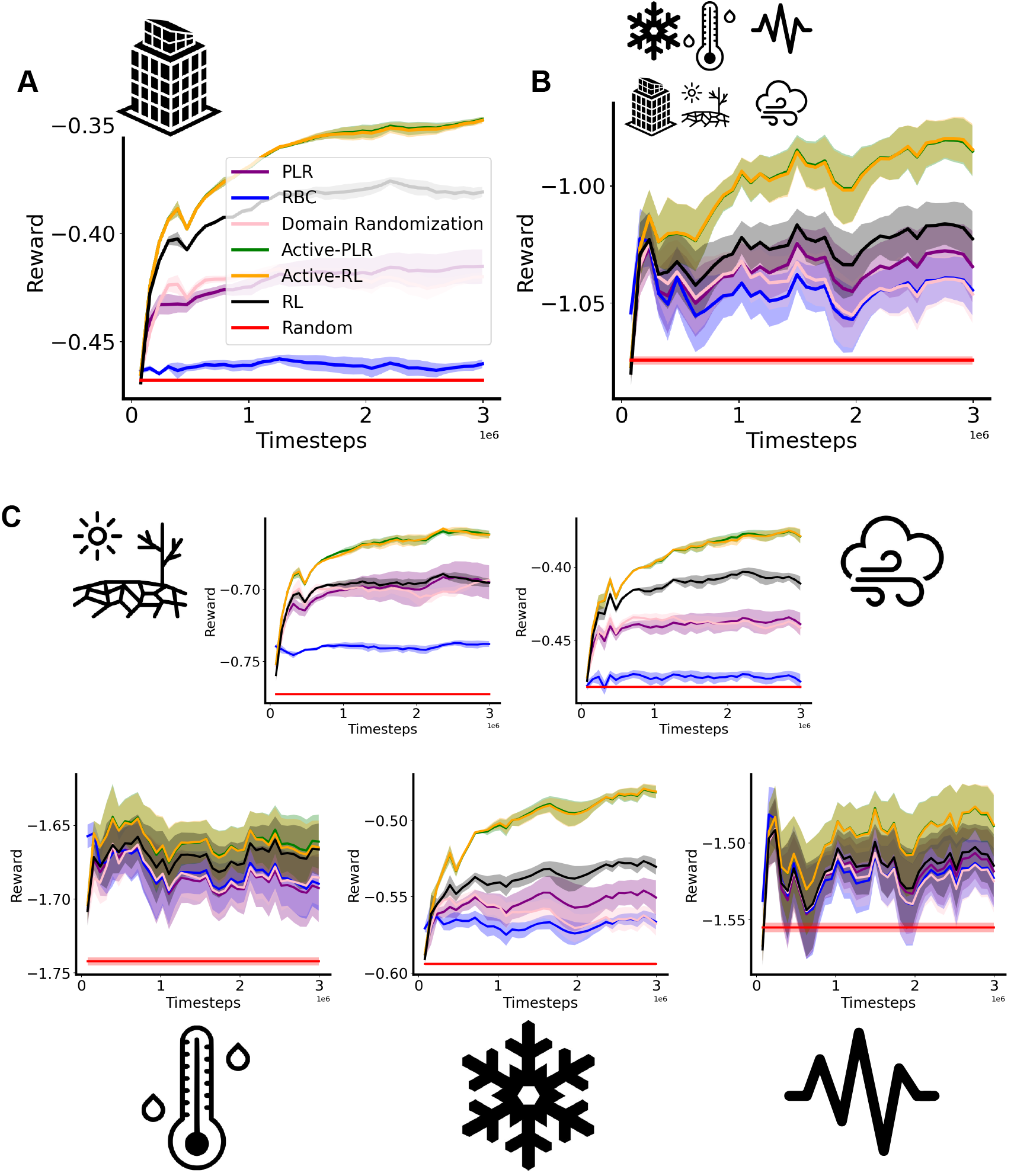}
    \caption{Performance of each algorithm on training an RL HVAC agent, tested on various weather patterns. ActiveRL and ActivePLR outperform all baselines. We report the standard error of the mean over 5 trials for each result. \textbf{A.} Average reward achieved by each algorithm on the base environment $\phi_0$ throughout training. \textbf{B.} Average reward achieved by each algorithm averaged over all 6 environments throughout training. \textbf{C.} Average reward achieved by each algorithm in each of the 5 extreme weather environments throughout training. Note that none of the algorithms were trained on these specific extreme weather environments.}
    \label{fig:results_big}
\end{figure*}
\begin{table*}[ht]
\centering
\caption{Bounds for Sinergym environment configuration variables}
\label{tab:bounds}
\begin{tabular}{>{\centering\arraybackslash}m{4cm} >{\centering\arraybackslash}m{4.5cm} >{\centering\arraybackslash}m{4.5cm}}
\toprule
\textbf{Weather Variable} & \textbf{Lower Bound} & \textbf{Upper Bound} \\
\midrule
Outdoor Air Temperature (\textdegree C) & $-31.05$ & $60.7$ \\
\midrule
Outdoor Air Relative Humidity ($\%$) & $3$ & $100$ \\
\midrule
Outdoor Wind Speed (m/s) & $0.0$ & $23.1$ \\
\midrule
Outdoor Wind Direction (\textdegree) & $0$ & $360$ \\
\midrule
Direct Solar Radiation Rate (W) & $0$ & $1033$ \\
\bottomrule
\end{tabular}
\end{table*}

\begin{table*}[ht]
\centering
\caption{Relevant hyperparameters}
\label{tab:hyperparameters}
\begin{tabular}{>{\centering\arraybackslash}m{2.5cm} >{\centering\arraybackslash}m{6.5cm}}
\toprule
\textbf{Algorithm} & \textbf{Hyperparameters} \\
PPO \tablefootnote{Note that all other algorithms share these hyperparameters} & lr$=0.00005$,  clip\_param$=0.3$, discount\_factor$=0.8$, $p_{dropout}=0.1$, \# of inner SGD steps$=40$\\
\midrule
ActiveRL & $\gamma=0.5$, $\eta=0.01$, $p_{dropout}=0.1$, $N=91$, $C=10$\\
\midrule
PLR & $\rho=0.045$, $\beta=0.0015$, $N_{PLR}=10$ \\
\midrule
ActivePLR \tablefootnote{We use the default parameters from \citet{jiang2021prioritized} for the PLR part of ActivePLR because our hyperparameter sweep showed that none of the PLR-specific hyperparameters made much of a difference for ActivePLR. Other hyperparameters are the same as ActiveRL}& $\rho=0.1$, $\beta=0.1$, $N_{PLR}=100$ \\
\bottomrule
\end{tabular}
\end{table*}

\begin{table*}[ht]
\centering
\caption{Hyperparameter sweep ranges}
\label{tab:sweep}
\begin{tabular}{>{\centering\arraybackslash}m{2.5cm} >{\centering\arraybackslash}m{8cm}}
\toprule
\textbf{Algorithm} & \textbf{Hyperparameters included in Sweep} \\
\midrule
PPO \tablefootnote{Note that all other algorithms share these hyperparameters} & lr$\in \{0.0005, 0.00005, 0.000005\}$,  clip\_param$\in \{0.1, 0.2, 0.3\}$, discount\_factor$\in \{0.8, 0.9, 0.99\}$, \# of inner SGD steps$\in{20, 30, 40}$\\
\midrule
ActiveRL & $\gamma \in \{0, 0.0005, 0.005, 0.05, 0.5\}$, $\eta \in [e^{-10}, 1]$, $p_{dropout} \in \{0.1, 0.25 ,0.5\}$, $N \in [1, 100]$ \\
\midrule
PLR & $\rho \in [e^{-8}, 1]$, $\beta \in [e^{-8}, 1]$, $N_{PLR} \in \{10, 50, 100, 200\}$ \\
\midrule
ActivePLR & $\gamma \in \{0, 0.0005, 0.005, 0.05, 0.5\}$, $\eta \in [e^{-10}, 1]$, $p_{dropout} \in \{0.1, 0.25 ,0.5\}$, $N \in [1, 100]$, $\rho \in [e^{-8}, 1]$, $\beta \in [e^{-8}$,  $N_{PLR} \in \{10, 50, 100, 200\}$ \\
\bottomrule
\end{tabular}
\end{table*}
\section{How Sinergym Resets Weather Conditions}
\label{sec:ou_details}
In order to simulate the outdoor weather, Sinergym takes as input a file that contains hourly measurements of each of the outdoor weather variables denoted with a '*' above, as well as several others. Originally, Sinergym added noise to the measured outdoor temperature through an Ornstein-Uhlenbeck (OU, \citealp{doob1942brownian}) process, to help prevent overfitting the RL agent to the static weather pattern. We modified Sinergym so that it could add this noise to the other weather variables denoted with a '*' as well. An OU process has three parameters: $\sigma$, $\mu$, and $\tau$. If we have a noise vector $x_t$, then 
\begin{equation}
\label{eq:OU}
    x_{t+1} = x_t + \Delta t * (-(x_t-\mu)/\tau) + \sigma * \sqrt{\frac{2}{\tau}}Z
\end{equation} where $Z \sim Normal(0, 1)$. so $\sigma$ controls the magnitude of the noise that is added, $\mu$ is the average value of the noise, and $\tau$ determines how quickly the noise reverts to the mean. Notably, if we have a recorded weather variable $w_t$, then adding the noise results in $\text{Mean}_{w_t+x_t}=\text{Mean}_{w_t} + \mu$. For each of our 5 weather variables, we estimate realistic values for $\sigma$ and $\tau$ by doing linear regression of the difference between that weather variable and its moving average. That is, if we assume our recorded weather variable $w_t$ was generated via adding noise generated through an OU process, then we can generate measurements $x_t = w_t - \text{MA}(w_t)$, where MA is the moving average. By applying linear regression onto the generated $x_t$'s, we can estimate values of $\sigma$ and $\tau$ that will generate weather with a similar amount of noise to real weather conditions. A detailed description of this linear regression process is detailed in \Cref{sec:us_setup}.

\section{Evaluating ActivePLR's Generalization to US Weather Conditions}
\label{sec:us_setup}
\subsection{Experiment Setup}
Since we handcrafted the extreme weather environments in \Cref{sec:extreme_setup}, it is possible that these environments are unrealistic. In order to properly assess the viability of our RL HVAC controller in a range of different weather scenarios, we constructed a dataset of 120 randomly sampled, recorded weather patterns from across the US. We deployed the HVAC controller for ActivePLR and each baseline in a building that simulated each of those 120 weather patterns. 

To construct the dataset of 120 weather patterns, we first scraped the EnergyPlus weather data website to get recorded weather patterns from across the US.\footnote{https://energyplus.net/weather} These were Typical Meteorological Year (TMY) weather patterns \cite{wilcox2008users} from NREL, which contain hourly meteorological information from specific weather stations over the course of 1 year (8760 hours).\footnote{e.g. temperature, humidity, wind, etc.} This meteorological information is specially collated from multiple historical recordings of the weather data in that location to present the range of weather phenomena that typically occur there, while still keeping to annual averages that are consistent with long term averages for that location. TMY weather data is used often for building simulations.

After we obtained a dataset of historical weather data recordings, we converted them into a realistic dataset of environment configuration parameters $\phi$. We modeled each weather variable in each weather pattern as a variation generated by an OU process(\Cref{eq:OU}) from the corresponding weather variable in the base environment configuration $\phi_0$. Formally, let us suppose we have some recorded weather variable $y \in \REAL^{8760}$, corresponding to the value of that weather variable for each hour in a year. We also have a recording corresponding to the base environment $\phi_0$, $y^0 \in \REAL^{8760}$. Since Sinergym takes the parameters of an OU process as its environment configuration, we model the difference $x_t = y_t - y^0_t$ as having been generated from an OU process, like in \Cref{eq:OU}. We rearrange the terms in \Cref{eq:OU} as:

\begin{equation}
    x_{t+1} = (1-\frac{\Delta t}{\tau})x_t + \frac{\mu \Delta t}{\tau} + \sigma * \sqrt{\frac{2}{\tau}}Z
\end{equation}

\begin{equation}
    x_{t+1} = m x_t + b + E
\end{equation} where $m=(1-\frac{\Delta t}{\tau})$, $b=\frac{\mu \Delta t}{\tau}$, and $E=\sigma * \sqrt{\frac{2}{\tau}}Z$. We can then run linear regression to find what parameters $m$ and $b$ estimate $x_{t+1}$ from $x$ while minimizing the error term $E$. Once we have estimated $m$ and $b$ with linear regression, we can compute the residual error $E = x_{t+1} - m x_t - b$ and compute the standard deviation of E as an estimate for $\sigma * \sqrt{\frac{2}{\tau}}$. Finally, we can estimate $\tau = \frac{\Delta t}{1-m}$, $\mu = \frac{b \tau}{\Delta t}$, $\sigma = \frac{\sqrt{Var(E)}}{\sqrt{\frac{2}{\tau}}}$.
We then repeat this process for each of the 120 US weather patterns, for each of the 5 weather variables that compose the environment configuration: outdoor humidity, air temperature, wind speed, wind direction, and solar irradiance. Thus we have a dataset $X \in \REAL^{120 \times 5 \times 3}$ of environment configurations that Sinergym can take in and simulate. \footnote{The final 3 dimension for $X$ comes from the fact that we have 3 variables $\mu, \sigma, \tau$ for the OU process for each weather variable.}\footnote{Note that the environment configuration variables that go into ActiveRL, RPLR, or DR $\phi \in \REAL^5$ are a subset of the full $\REAL^{5 \times 3}$ environment configuration that can be provided to Sinergym, as $\phi$ only contains the offset parameters $\mu$.}

\textbf{Our hypothesis is that by conducting an uncertainty-driven environment exploration that is constrained to realistic environments, ActivePLR will be able to generalize to different weather patterns across the US better than the baseline methods.} In particular, our hypothesis for DR and RPLR is that they will end up training the RL algorithm to focus on performing well in unrealistic environments, and cause performance to degrade on this set of more realistic environments.

\subsection{ActivePLR Generalizes to US Weather Conditions}
\begin{figure*}[h]
    \centering
    \includegraphics[width=0.8\textwidth]{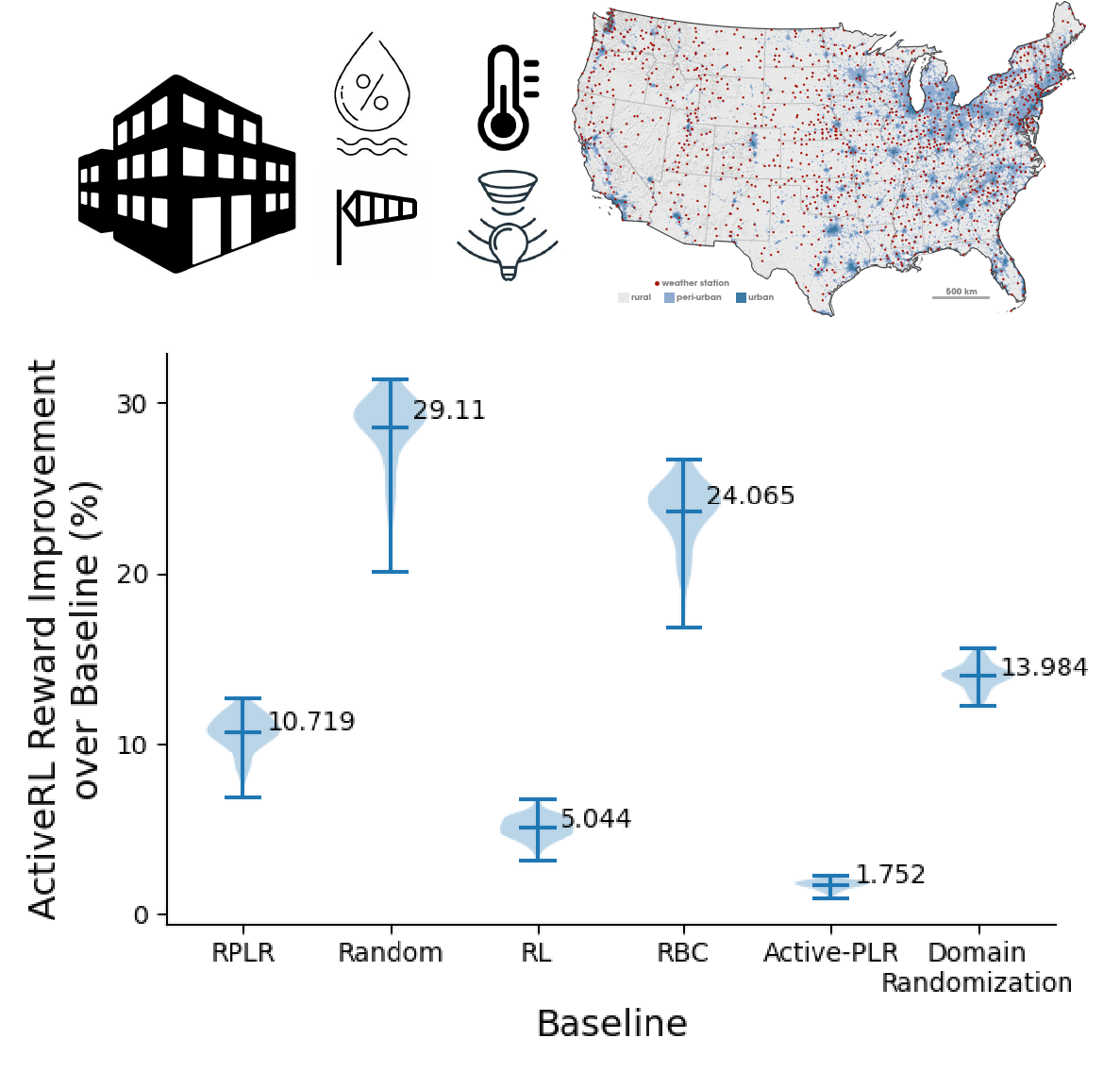}
    \caption{Improvement in reward achieved by using ActivePLR instead of each baseline, on 120 randomly sampled weather patterns from across the US. A higher number here indicates that ActivePLR performs well in comparison to that baseline. On the other hand, a higher number indicates the baseline performs poorly.}
    \label{fig:us_results}
\end{figure*}

\label{sec:us_results}
Although the ActivePLR agent seems to perform well in the handcrafted extreme weather conditions from \Cref{sec:extreme_results}, it is possible that these environments are unrealistic. In order to properly assess the viability of our RL HVAC controller in a range of different weather scenarios, we deployed the HVAC controller for ActivePLR and each baseline in a building that simulated each of those 120 different weather patterns sampled from across the US.

On all 120 environments, ActiveRL achieves a higher reward than every baseline, showing that our uncertainty-driven UED approach can train an RL HVAC agent that is more robust to realistic extreme weather patterns than any of our baselines. Since ActiveRL outperforms all the baselines on every environment, we visualize how much better the reward achieved by ActiveRL is relative to each baseline in each of the 120 different weather patterns in \Cref{fig:us_results}. The median improvement of ActiveRL over the RBC is $24\%$. Over vanilla RL, it is $5\%$. 

Interestingly, there is a very small improvement using ActiveRL over ActivePLR; this, combined with the similar performance between ActiveRL and ActivePLR from \Cref{sec:extreme_results} suggests that ActiveRL and ActivePLR have very similar behavior. One possible reason is that there may be some attractive local (or global) optimum that both algorithms fall into, resulting in them appearing to have similar performance. Another possible reason is that since PLR tries to sample environment configurations that result in high value loss, and ActiveRL tries to sample environment configurations that result in high value uncertainty, PLR and ActiveRL actually optimize for very similar objectives. Thus ActiveRL and ActivePLR end up having similar behavior, at least with respect to their responses to weather. One advantage that ActiveRL has in optimizing for uncertainty rather than value loss is that it can be used to identify novel environments to learn from rather than having to sample from old ones, or worry about the staleness of the value loss estimates of the environments in the replay buffer.

There is a significant difference between both DR and RPLR, and vanilla RL. Both UED methods seem to perform poorly compared to vanilla RL. This seems to indicate that randomly sampling environment configuration parameters results in environments that are very unrealistic, causing poor generalization performance compared to the vanilla RL algorithm or ActiveRL. Although RPLR has the ability to control what environments in its replay buffer are sampled, its replay buffer is still populated through the same uniform random sampling process as used in DR, resulting in training on unrealistic environments.

\section{Ablations Exploring Components of ActivePLR}
\label{sec:ablation_setup}

In order to further understand the driving factors behind the performance of ActivePLR, we conducted two ablation experiments. First, we explored the impact of the $\gamma$ soft constraint term. Second, we explored the impact of the learning rate $\eta$ on the performance of the algorithm. To better isolate the impact of each parameter without the added complexity of the PLR replay buffer, we conduct these ablation experiments on ActiveRL (which is just ActivePLR without the replay buffer).

\subsection{Constraints}
First, we explore the necessity of constraining ActiveRL from generating environment configurations $\phi$ that are too far away from $\phi_0$. $\gamma$ shows up in \Cref{eq:soft}, as a coefficient that regulates how much the distance of the generated environment configuration $\phi$ from the base environment $\phi_0$ contributes to the objective function of ActiveRL. As $\gamma$ increases, the algorithm is encouraged to generate values of $\phi$ that are closer to $\phi_0$. We varied $\gamma$ between four different values \{0, 0.005, 0.05, 0.5\} while keeping the other hyperparameters the same as our other experiments to better understand how the algorithm performs under different constraint strengths.

\textbf{Our hypothesis with this experiment was that there would be a tradeoff between performance in the extreme environments, and performance in the base environment $\phi_0$ that would be modulated by $\gamma$.}

\subsection{Learning Rate}
The learning rate $\eta$ determines the step-size used by the Adam optimizer when ActiveRL is conducting gradient ascent on $\phi$. The smaller $\eta$ is, the more fine-grained the search for an uncertain environment configuration becomes. We mainly explore this hyperparameter to assess how sensitive ActiveRL is to the user's choice of $\eta$; if there are many values of $\eta$ that yield optimal performance, then ActiveRL becomes much easier to use for other problems. We varied $\eta$ between five different values $\{0.0001, 0.001, 0.01, 0.1, 1.0\}$ while keeping the other hyperparameters the same as our other experiments. 

\textbf{Our hypothesis for this experiment was that there would be some optimal value of $\eta$ that yielded the best performance by striking the perfect balance between being large enough to avoid local minima, and being small enough to actually converge.}
\subsection{Results of Exploring Components of ActiveRL}
\label{sec:ablation}
\begin{figure*}
    \centering
    \includegraphics[width=\textwidth]{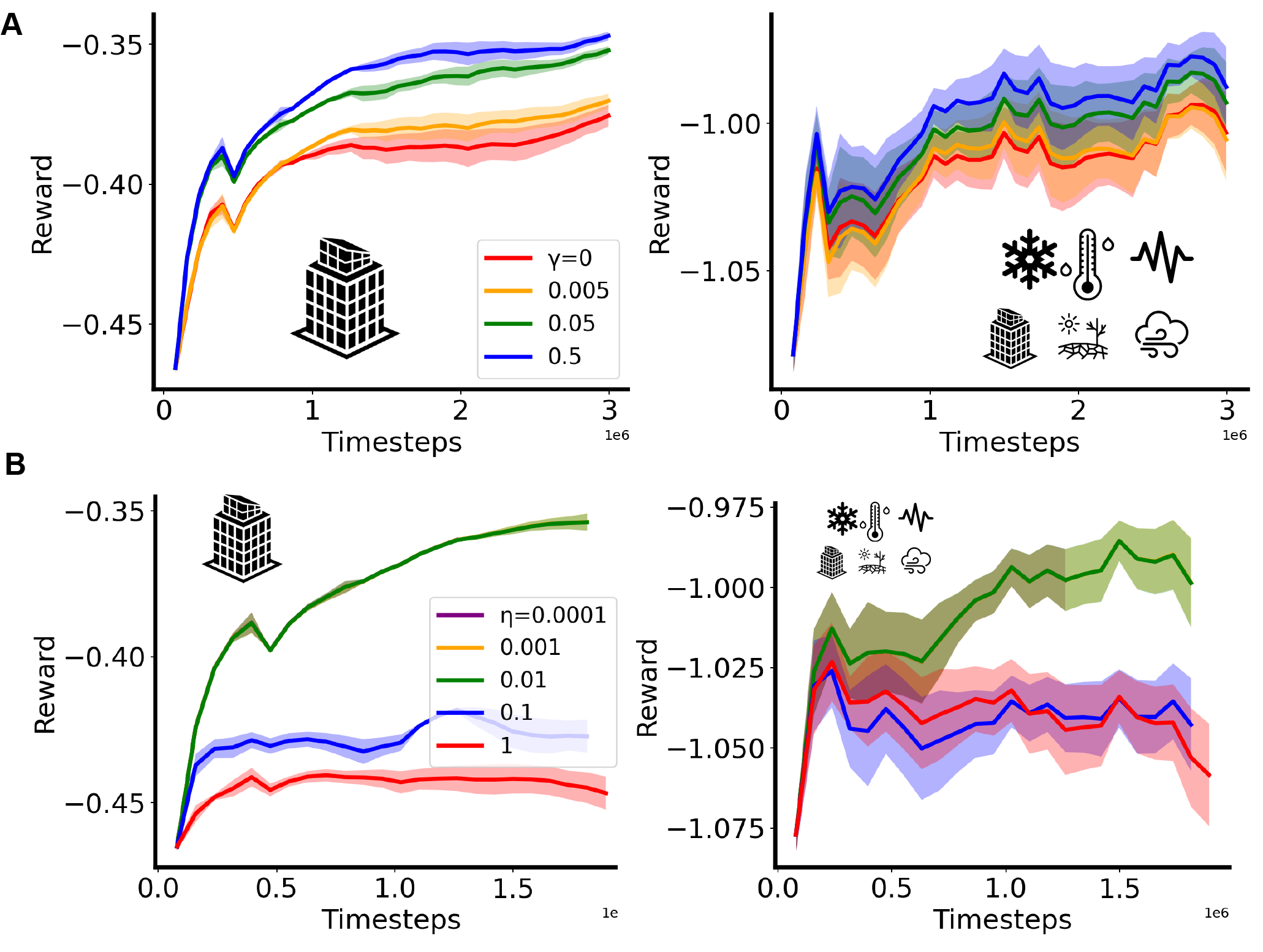}
    \caption{\textbf{Ablation} \textbf{A.} We explore how the $\gamma$ regularization parameter affects the performance of ActiveRL. Higher values of $\gamma$ mean that ActiveRL is forced to propose training environment configurations $\phi$ that are close to the base environment $\phi_0$. The left graph shows the average reward obtained throughout training on the base environment $\phi_0$. The right graph shows the reward averaged over $\phi_0$ and the 5 extreme weather environments. ActiveRL seems quite sensitive to $\gamma$. \textbf{B.} Similarly, we explore how the $\eta$ learning rate parameter affects ActiveRL. $\eta$ is the learning rate used by ActiveRL to conduct gradient ascent on $\phi$. Higher values of $\eta$ means a coarser-grained search for the $\phi$ that maximizes the agent's uncertainty. ActiveRL seems insensitive to $\eta$ once it gets small enough ($<0.1$).}
    \label{fig:ablation}
\end{figure*}
In order to assess what factors contribute to the performance of ActiveRL, we ran some ablation experiments that show how ActiveRL changes as certain hyperparameters change. We look at the $\gamma$ hyperparameter that controls the strength of the soft constraint on ActiveRL's environment design process, and the $\eta$ parameter that controls the step-size of the optimization procedure used to generate new environments. The results of changing these two parameters can be seen in \Cref{fig:ablation}, where panel A corresponds to testing different values of $\gamma$ and panel B corresponds to different values of $\eta$.

Contrary to our original hypothesis that there would be some tradeoff between realism and robustness modulated by $\gamma$, we actually found that having a relatively high value of $\gamma$ contributes to good performance in both the base environment and the extreme environments. There was a clear pattern that larger values of $\gamma$ correlated well with better performance. This may be because ActiveRL will generate more unrealistic environment configurations $\phi$ with weaker regularization that are not similar enough to $\phi_0$ and the extreme environments to aid performance in those settings.

We found that smaller values of $\eta$ helped performance across all environments, but decreasing it below $0.001$ did not change the agent's learning trajectory at all. It is possible that having a large $\eta$ results in an unstable gradient ascent process which is unable to successfully find a $\phi$ that maximizes the agent's uncertainty.
\end{document}